\documentclass{article}

\usepackage[nonatbib,preprint]{neurips_2020}

\usepackage[utf8]{inputenc} % allow utf-8 input
\usepackage[T1]{fontenc}    % use 8-bit T1 fonts
\usepackage{hyperref}       % hyperlinks
\usepackage{url}            % simple URL typesetting
\usepackage{booktabs}       % professional-quality tables
\usepackage{amsfonts}       % blackboard math symbols
\usepackage{nicefrac}       % compact symbols for 1/2, etc.
\usepackage{microtype}      % microtypography

\usepackage{subfigure}
\usepackage{graphicx}
\usepackage{amsmath,amssymb}
\usepackage{algorithm}
\usepackage{algorithmicx}
\usepackage[noend]{algpseudocode}
\usepackage{latexsym}
\usepackage{xspace}

\begin{document}

% shorthands for naming conventions
\newtheorem{problem}{Problem}
% macro symbol definitions

%algorithm
%\renewcommand{\algorithmicrequire}{\textbf{Input:}}
%\renewcommand{\algorithmicensure}{\textbf{Output:}}

%NTR3
\newcommand{\nameNTR}{Linear Tensor Projection for Nonlinear Prediction\xspace}
\newcommand{\NTR}{LTPNP\xspace}
\newcommand{\nameProposed}{Tensor Reconstruction based Interpretable Prediction\xspace}
\newcommand{\Proposed}{TRIP\xspace}
\newcommand{\nameLRC}{local regression coefficients\xspace}
\newcommand{\LRC}{LRC\xspace}
\newcommand{\PCANN}{PCANN\xspace}
\newcommand{\LDANN}{LDANN\xspace}
\newcommand{\TuckerNN}{TuckerNN\xspace}
\newcommand{\nameSpiral}{Spiral\xspace}
\newcommand{\Spiral}{Spiral\xspace}
\newcommand{\nameRandomone}{Random-1st\xspace}
\newcommand{\Randomone}{Rand1\xspace}
\newcommand{\nameConnectionist}{Connectionist Bench (Sonar, Mines vs. Rocks)\xspace}
\newcommand{\Connectionist}{Connectionist\xspace}
\newcommand{\namequality}{Quality Assessment of Digital Colposcopies\xspace}
\newcommand{\quality}{Colposcopies\xspace}
\newcommand{\nameAlizadeh}{Z-Alizadeh Sani\xspace}
\newcommand{\Alizadeh}{Alizadeh\xspace}
\newcommand{\nameSports}{Sports articles for objectivity analysis\xspace}
\newcommand{\Sports}{Sports\xspace}
\newcommand{\nameSECOM}{SECOM\xspace}
\newcommand{\SECOM}{SECOM\xspace}
\newcommand{\nameOzone}{Ozone Level Detection\xspace}
\newcommand{\Ozone}{Ozone\xspace}
\newcommand{\nameSpambase}{Spambase\xspace}
\newcommand{\Spambase}{Spambase\xspace}
\newcommand{\nametheo}{First-order theorem proving\xspace}
\newcommand{\theo}{Theorem\xspace}
\newcommand{\namemusk}{Musk (Version 2)\xspace}
\newcommand{\musk}{Musk\xspace}
\newcommand{\nameEpileptic}{Epileptic Seizure Recognition\xspace}
\newcommand{\Epileptic}{Epileptic\xspace}
\newcommand{\nameIris}{Iris\xspace}
\newcommand{\Iris}{Iris\xspace}
\newcommand{\nameWine}{Wine\xspace}
\newcommand{\Wine}{Wine\xspace}
\newcommand{\nameGeneExpression}{Gene Expression\xspace}
\newcommand{\GeneExpression}{Genes\xspace}
\newcommand{\nameRandomtwo}{Random-2nd\xspace}
\newcommand{\Randomtwo}{Rand2\xspace}
\newcommand{\nameRandomthree}{Random-3rd\xspace}
\newcommand{\Randomthree}{Rand3\xspace}
\newcommand{\nameVC}{varying coefficients\xspace}
\newcommand{\VC}{VC\xspace}
\newcommand{\nameTG}{target genes\xspace}
\newcommand{\TG}{TG\xspace}
\newcommand{\nameRG}{regulator genes\xspace}
\newcommand{\RG}{RG\xspace}

\title{Linear Tensor Projection Revealing Nonlinearity}

\author{%
  Koji Maruhashi ~\thanks{Fujitsu Laboratories Ltd., 4-1-1 Kamikodanaka, Nakahara-ku, Kawasaki, Kanagawa, Japan}\\
  \texttt{maruhashi.koji@jp.fujitsu.com} \\
  \And
  Heewon Park ~\thanks{M\&D Data Science Center, Tokyo Medical and Dental University, 1-5-45 Yushima, Bunkyo-ku, Tokyo, Japan}\\
  \texttt{hwpark.dsc@tmd.ac.jp} \\
  \AND
  Rui Yamaguchi
~\thanks{Division of Cancer Systems Biology, Aichi Cancer Center Research Institute, 1-1 Kanokoden, Chikusa-ku, Nagoya, Aichi, Japan}
~\thanks{Division of Cancer Informatics, Nagoya University Graduate School of Medicine, 65 Tsurumai-cho, Showa-ku, Nagoya, Aichi, Japan}
~\thanks{Human Genome Center, The Institute of Medical Science, The University of Tokyo, 4-6-1 Shirokane-dai, Minato-ku, Tokyo, Japan}\\
  \texttt{r.yamaguchi@aichi-cc.jp} \\
  \And
  Satoru Miyano
~\footnotemark[2]
~\footnotemark[5]\\
  \texttt{miyano@hgc.jp} \\
}

\maketitle

\begin{abstract}
Dimensionality reduction is an effective method for learning high-dimensional data, which can provide better understanding of decision boundaries in human-readable low-dimensional subspace. Linear methods, such as principal component analysis and linear discriminant analysis, make it possible to capture the correlation between many variables; however, there is no guarantee that the correlations that are important in predicting data can be captured.
 Moreover, if the decision boundary has strong nonlinearity, the guarantee becomes increasingly difficult. This problem is exacerbated when the data are matrices or tensors that represent relationships between variables. We propose a learning method that searches for a subspace that maximizes the prediction accuracy while retaining as much of the original data information as possible, even if the prediction model in the subspace has strong nonlinearity. This makes it easier to interpret the mechanism of the group of variables behind the prediction problem that the user wants to know. We show the effectiveness of our method by applying it to various types of data including matrices and tensors.

\end{abstract}

\section{Introduction}
\label{sec.intro}
\begin{figure}[ttt]
\begin{center}
\includegraphics[width=0.98\columnwidth]{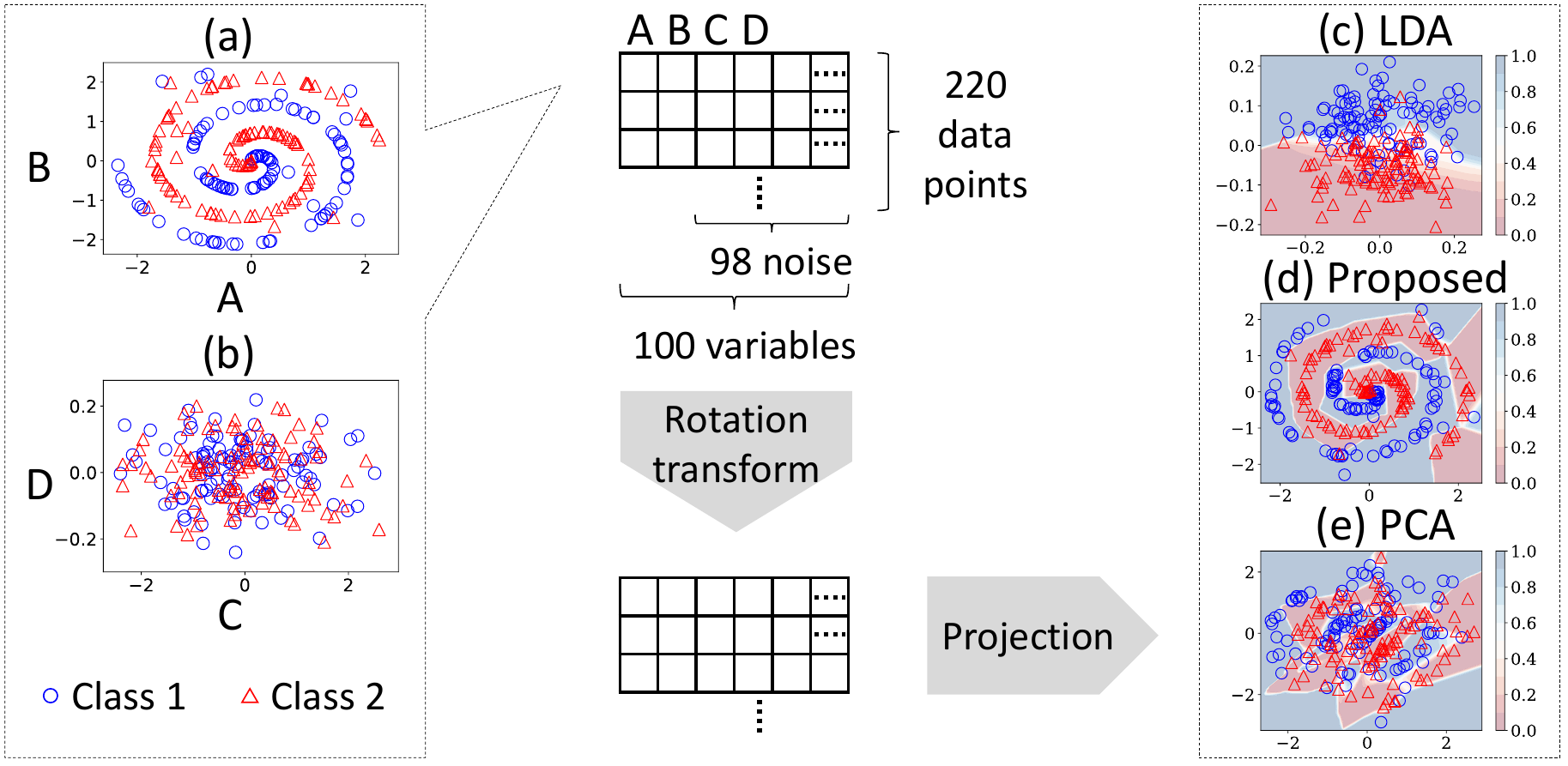}
\caption{
Two-dimensional spiral distribution (a) buried in $98$-dimensional noise (b) by a rotation transform are extracted by proposed method (d), whereas LDA overfits the noise (c) and PCA fails to extract the spiral distribution (e). Contours represent the softmax values for class $1$ of the neural network applied to the projection.\label{fig.intro.galaxy}}
\end{center}
\end{figure}

Dimensionality reduction is effective for learning high-dimensional data.
An important effect is avoiding overfitting by revealing a few essential parameters.
Another important effect is providing better understanding of decision boundaries in human-readable low dimensional subspace.
We can understand the reason of prediction by analyzing the positional relationship between data points and decision boundaries.
Several dimensionality reduction methods, such as t-SNE~\cite{Maaten08}, kernel PCA~\cite{Barshan11,Hosseini19,Scholkopf98}, and Isomap~\cite{tenenbaum_global_2000}, nonlinearly reduce dimensionality; however, the relationship between original variables and projections is not straightforward and difficult to understand in many cases.
On the other hand, it may be difficult for variable selection methods such as the celebrated LASSO approach~\cite{tibshirani96} to explain the phenomenon such as biological mechanism that should be described as the interaction between the sets of the variables that slightly correlate with each other.
Linear methods without variable selection, such as principal component analysis (PCA) and linear discriminant analysis (LDA), are desirable from the interpretation point of view because we can easily understand the correlation between many variables in the projected low-dimensional subspace.
We can apply any prediction model to the projected data point; however, there is no guarantee that the important features for prediction are kept in the projected subspace.
To highlight these issues, we give a vivid example using artificial data of a $2$-class classification task with $100$-dimensional variables.
These data are created by embedding the two-dimensional spiral distribution (Fig.~\ref{fig.intro.galaxy}(a)) in a $100$-dimensional space by applying $98$-dimensional noise (Fig.~\ref{fig.intro.galaxy}(b)) and performing rotation transformation (see details in Section~\ref{sec.exp}).
Figure~\ref{fig.intro.galaxy}(c) shows a distribution of the data points projected to two-dimensional subspace using LDA, a supervised method.
However, it overfit the noise and the spiral distribution cannot be observed at all.
By contrast, PCA seems not to overfit the noise; however, it is difficult to recognize the spiral distribution (Fig.~\ref{fig.intro.galaxy}(e)).
Thus, it is difficult for linear projection methods to deal with strong nonlinearities in high-dimensional space.
In this paper, we propose a learning method of searching for a subspace that maximizes the prediction accuracy while retaining as much of the original data variance as possible to optimize the projection matrix ${\bf C} \in \mathbb{R}^{I \times J}$ to minimize {\it regularized-by-reconstruction loss} such as
\begin{equation}\label{eq.intro.regrecon}
E = \frac{1}{N} \sum_{n=1}^{N} L({\bf C}^T{\bf x}_n, {\bf \theta}; y_n) + \lambda \left\| {\bf x}_n - {\bf C}{\bf C}^T{\bf x}_n \right\|_2^2 \quad s.t. \quad {\bf C}^T {\bf C} = {\bf I},
\end{equation}
where ${\bf x}_n \in \mathbb{R}^I$ is the $n$-th data point, $L(\cdot)$ is a loss of the prediction model to predict a scalar response $y_n$ under the subspace, ${\bf \theta} \in \mathbb{R}^{H}$ are the parameters of the prediction model, ${\bf I}$ is an identity matrix, and $\lambda > 0$ is the regularization parameter.
This strategy meets two requirements simultaneously.
First, the subspace keeps the variance of the data points as large as possible by minimizing the reconstruction errors $\left\| {\bf x}_n - {\bf C}{\bf C}^T{\bf x}_n \right\|_2^2$, the same as with PCA.
Second, $y_n$ can be predicted as accurately as possible by minimizing $L(\cdot)$, even if the prediction model requires strong nonlinearity.
We can successfully find the spiral distribution, as shown in Fig.~\ref{fig.intro.galaxy}(d), by calculating an optimal ${\bf C}$ satisfying Eq.~\ref{eq.intro.regrecon}.

The issues described above also apply to the tensor data.
Several tensor-regression methods have been proposed for the data represented by matrices or tensors that describe the relationships between variables~\cite{guhaniyogi_bayesian_2017,he_boosted_2018,imaizumi_doubly_2016,yu_learning_2016}.
However, these methods cannot cope with strong nonlinearities.
On the other hand, several methods using the kernel method~\cite{he_kernelized_2017,Tao07} lose the correlation between many variables of the original data.
We extend Eq.~\ref{eq.intro.regrecon} to tensor expressions so that they can be applied to tensor data.

To the best of our knowledge, this is the first study to propose a method that allows searching for a linear projection without variable selection that maintains the maximum data variation and allows high-accuracy prediction even with strong nonlinearity, including matrix and tensor data.
We report the results of evaluating the proposed method using artificial data and real-world data including higher-order data.
We empirically show that our method can classify high-dimensional data with high accuracy while having high interpretability.

\section{Related Works}
\label{sec.related}
PCA is the most well-known unsupervised dimensionality-reduction method that represents a high-dimensional space by a few orthogonal variables that retain large variance of the data~\cite{jolliffe_principal_2002}.
Several methods have been proposed to find low-dimensional subspace in a supervised manner, such as LDA and partial least squares (PLS)~\cite{Martinez01,Rosipal06}.
However, the distribution of samples in each class is strongly assumed as a normal distribution and cannot be applied to datasets with strong nonlinearity as shown in Fig.~\ref{fig.intro.galaxy}(a).
Several dimensionality-reduction methods, such as t-SNE~\cite{Maaten08}, kernel PCA~\cite{Barshan11,Hosseini19,Scholkopf98}, and Isomap~\cite{tenenbaum_global_2000}, nonlinearly reduces dimensionality; however, the relationship between original variables and projections is not straightforward and difficult to understand in many cases.

Tucker decomposition~\cite{Kolda09} is a well-known tensor decomposition using unsupervised methods.
Several studies on tensor regression task have been reported~\cite{guhaniyogi_bayesian_2017,he_boosted_2018,imaizumi_doubly_2016,yu_learning_2016}.
Most of these methods reduce the number of parameters by expressing the partial regression coefficients in a low rank and can be regarded as the multi-linear projection into lower-dimensional tensor subspace.
However, these methods cannot cope with the nonlinearity inherent in the data.
Kernel methods have been applied to deal with nonlinearities~\cite{he_kernelized_2017,Tao07}; however, they lose the correlation between many variables.

Prediction interpretation and justification is one of the most important topics in recent machine learning research~\cite{Adadi18,Lipton18}.
One effective approach is to produce {\em interpretable models}.
Ribeiro et al. introduced a model-agnostic approach that provides explanation to blackbox models by learning interpretable models~\cite{Ribeiro16}.
Our method allows us to better understand the interpretable model by visualizing it along with the decision boundaries in a lower human-readable dimension.

\section{Preliminaries}
\label{sec.prelim}
\subsection{Notations}

Scalars are denoted as lowercase letters ($a$), vectors as boldface lowercase letters (${\bf a}$), matrices as boldface capital letters (${\bf A}$), and tensors as boldface Euler script letters (${\bf \mathcal{A}}$).
A transverse matrix of ${\bf A}$ is ${\bf A}^T$.
A matrix ${\bf I}$ is an identity matrix, and ${\bf 1}$, ${\bf 0}$ are square matrices in which all values are $1$, $0$.
The square root of the sum of the squared values in tensor ${\bf \mathcal{A}}$ is denoted as $\|{\bf \mathcal{A}}\|_2$.
We refer to the dimensionality of a tensor as {\em order} and to each dimension as a {\em mode}.
The inner product is denoted as $\langle \cdot , \cdot \rangle$.
The outer product of vectors is denoted as ${\bf a}^{(1)} \circ \cdots \circ {\bf a}^{(K)}$, where the $(i_1\ldots i_K)$-th element is $a_{i_1}^{(1)} \ldots a_{i_K}^{(K)}$ by using $a_{i_k}^{(k)}$ as the $i_k$-th element of ${\bf a}^{(k)}$.
The element-wise multiplication and division are denoted as $\ast$ and $\oslash$, respectively.
A Khatri-Rao product between ${\bf A} = \left[{\bf a}_1 \cdots {\bf a}_R \right]$ of size $I_1 \times R$ and ${\bf B} = \left[{\bf b}_1 \cdots {\bf b}_R \right]$ of size $I_2 \times R$ is a matrix ${\bf A} \odot {\bf B} = \left[{\bf a}_1 \otimes {\bf b}_1 \cdots {\bf a}_R \otimes {\bf b}_R \right]$ of size $I_1I_2 \times R$ by using a Kronecker product $\otimes$.
The vectorization of a tensor ${\bf \mathcal{A}}$ creates a vector $vec({\bf \mathcal{A}})$.
The mode-$k$ matricization of a tensor ${\bf \mathcal{A}}$ of size $I_1 \times \ldots \times I_K$ creates a matrix ${\bf \mathcal{A}}_{(k)} = [ vec({\bf \mathcal{A}}_{1}^{(k)}) \cdots vec({\bf \mathcal{A}}_{I_k}^{(k)}) ]$ of size $I_1\ldots I_{k-1}I_{k+1}\ldots I_K \times I_k$, where ${\bf \mathcal{A}}_{l}^{(k)}$ is a tensor of the $l$-th slice of ${\bf \mathcal{A}}$ for the $k$-th mode.
A $k$-mode product of a tensor ${\bf \mathcal{A}}$ with ${\bf W}$ of size $I_k \times I'_k$ is a tensor ${\bf \mathcal{A}}' = {\bf \mathcal{A}} \times_k {\bf W}$ of size $I_1 \times \ldots$ $\times I_{k-1} \times I'_{k} \times I_{k+1} \times \ldots \times I_K$, where ${\bf \mathcal{A}}'_{(k)} = {\bf \mathcal{A}}_{(k)} {\bf W}$.
In case ${\bf W}$ is a vector ${\bf w}$ of length $I_k$, we remove the mode and resize ${\bf \mathcal{A}}'$ to $I_1 \times \ldots \times I_{k-1} \times I_{k+1} \times \ldots \times I_K$.
We denote ${\bf \mathcal{A}} \times_1 {\bf W}_1 \ldots \times_K {\bf W}_K$ as ${\bf \mathcal{A}} \prod_{k} \times_k {\bf W}_k$.
${\bf \mathcal{A}} \odot_{k} {\bf W}$ is a tensor of size $I_1 \times \ldots \times I_K \times J$ with ${\bf W}$ of size $I_k \times J$, and $({\bf \mathcal{A}} \odot_{k} {\bf W})_{(k)} = {\bf \mathcal{A}}_{(k)} \odot {\bf W}^T$.
We denote ${\bf \mathcal{A}} \odot_1 {\bf W}_1 \ldots \odot_1 {\bf W}_K$ as ${\bf \mathcal{A}} \prod_{k} \odot_1 {\bf W}_k$.
A vector $\frac{\partial E}{\partial {\bf a}}$ and matrix $\frac{\partial E}{\partial {\bf A}}$ are differentials of $E$ over the elements of ${\bf a}$ and ${\bf A}$, respectively.

\subsection{Problem Definition}

As we mention in Section~\ref{sec.intro}, we extend our problem to tensor expression.
Among several ways to project a tensor ${\bf \mathcal{X}}_n$ of size $I_1 \times \cdots \times I_K$ to a tensor $\overline{{\bf \mathcal{X}}_n}$ of size $J_1 \times \cdots \times J_K$, we propose projecting ${\bf \mathcal{X}}_n$ by the $k$-mode product of each mode using a projection matrix ${\bf C}^{(k)} \in \mathbb{R}^{I_k \times J_k}$, {\it i.e.}, $\overline{{\bf \mathcal{X}}_n} = {\bf \mathcal{X}}_n \prod_k \times_k {\bf C}^{(k)}$, from the interpretation point of view.
Thus, our problem can be defined as follows:

\begin{problem}[\nameNTR(\NTR)]
Given a set of training samples $\{({\bf \mathcal{X}}_n, y_n)\}_{n=1}^{N}$ where ${\bf \mathcal{X}}_n$ is a $K$th-order tensor of size $I_1 \times \cdots \times I_K$ and $y_n$ is a scalar response, how can we learn the projection matrices $\{{\bf C}^{(k)} \in \mathbb{R}^{I_k \times J_k}\}_{k=1}^{K}$ to project ${\bf \mathcal{X}}_n$ into $\overline{{\bf \mathcal{X}}_n} = {\bf \mathcal{X}}_n \prod_k \times_k {\bf C}^{(k)}$ along with the prediction model $y'_n = pred(\overline{{\bf \mathcal{X}}_n};{\bf \theta})$ with strong nonlinearity that can accurately predict the response $y_n$ based on the projected tensor $\overline{{\bf \mathcal{X}}_n}$, where ${\bf \theta}$ are parameters of the prediction model?
\end{problem}

\section{Proposed Method}
\label{sec.proposed}
We propose a practical method of solving the \NTR problem.
We will refer to this method as {\it \nameProposed (\Proposed)}.

\subsection{Prediction}

As we explain later, we propose learning the projection matrices ${\bf C}^{(k)}$ and the prediction model by using stochastic gradient descent (SGD).
Typical prediction models that can be trained by using SGD include neural networks (NNs) and linear/softmax regression.
Such a prediction model can be written as
\begin{equation}\label{eq.proposed.model}
pred \left(\overline{{\bf \mathcal{X}}_n}; {\bf \theta} \right) = pred^{\dagger} \left( {\bf r}_n; {\bf \theta}' \right),
\end{equation}
where ${\bf r}_n = {\bf \mathcal{W}}_{(1)}^T vec \left( \overline{{\bf \mathcal{X}}_n} \right)$, ${\bf \mathcal{W}}$ is a weight tensor of size $M \times J_1 \times \cdots \times J_K$, $pred^{\dagger} (\cdot)$ is a prediction model obtained by removing ${\bf \mathcal{W}}$ from $pred(\cdot)$ with input of size $M$ vector, and ${\bf \theta}'$ are the remaining parameters other than ${\bf \mathcal{W}}$.
If the prediction model is linear/softmax regression, ${\bf \mathcal{W}}$ is the partial regression coefficients.
Moreover, if the prediction model is an NN, ${\bf \mathcal{W}}$ is the parameters between $\overline{{\bf \mathcal{X}}_n}$ and the first layer.
In any case, the number of elements of ${\bf \mathcal{W}}$ increases by the multiplication of the number of modes, so that over-fitting easily occurs.
As a solution to this problem, we set ${\bf \mathcal{W}}$ to be a low-rank tensor as 
\begin{equation}\label{eq.proposed.WG}
{\bf \mathcal{W}} = {\bf G}^{(1)T} \odot_1 \cdots \odot_1 {\bf G}^{(K)T}
\end{equation}
by using ${\bf G}^{(k)} \in \mathbb{R}^{J_k \times M}$, so that the number of parameters can be reduced.

\subsection{Training}

We show a method of learning $\{{\bf C}^{(k)}\}_{k=1}^{K}$, $\{{\bf G}^{(k)}\}_{k=1}^{K}$, and ${\bf \theta}'$, using SGD.
We extend the {\it regularized-by-reconstruction loss} (Eq.~\ref{eq.intro.regrecon}) to tensor expression as
\begin{equation}\label{eq.proposed.regrecontensor}
E = \frac{1}{N} \sum_{n=1}^{N} L(\overline{{\bf \mathcal{X}}_n}, {\bf \theta}; y_n) + \lambda \left\| {\bf \mathcal{X}}_n - \overline{{\bf \mathcal{X}}_n} \prod_k \times_k {\bf C}^{(k)T} \right\|_2^2 \quad s.t. \quad {\bf C}^{(k)T} {\bf C}^{(k)} = {\bf I},
\end{equation}
where $\lambda > 0$ is the regularization parameter.
It is necessary to satisfy the orthonormal condition ${\bf C}^{(k)T} {\bf C}^{(k)} = {\bf I}$.
We propose a similar idea as ~\cite{Maruhashi18}, that uses ${\bf Z}^{(k)}$ of the same size as ${\bf C}^{(k)}$ as latent variables and calculate ${\bf C}^{(k)}$ as a matrix that satisfies the orthonormal conditions derived from ${\bf Z}^{(k)}$.
That is,
\begin{equation}\label{eq.proposed.svd}
{\bf Z}^{(k)} = {\bf P}^{(k)}{\bf S}^{(k)}{\bf Q}^{(k)T}
\end{equation}
is calculated by singular value decomposition (SVD) and set as
\begin{equation}\label{eq.proposed.calcC}
{\bf C}^{(k)} = {\bf P}^{(k)}{\bf Q}^{(k)T},
\end{equation}
where ${\bf P}^{(k)}$ and ${\bf Q}^{(k)}$ are matrices having left and right singular vectors as column vectors, and ${\bf S}^{(k)}$ is a diagonal matrix having singular values as diagonal components.
Our idea is to calculate $\frac{\partial {\bf C}^{(k)}}{\partial z_{ij}^{(k)}}$ as a derivative of ${\bf C}^{(k)}$ by each element $z_{ij}^{(k)}$ of ${\bf Z}^{(k)}$ and calculate $\frac{\partial E}{\partial z_{ij}^{(k)}} = \left\langle \frac{\partial E}{\partial {\bf C}^{(k)}}, \frac{\partial {\bf C}^{(k)}}{\partial z_{ij}^{(k)}} \right\rangle$ to update ${\bf C}^{(k)}$ by updating ${\bf Z}^{(k)}$.
As a result, we can calculate
\begin{equation}\label{eq.proposed.calcgradZ}
\frac{\partial E}{\partial {\bf Z}^{(k)}} = f \left( \frac{\partial E}{\partial {\bf C}^{(k)}}, {\bf Z}^{(k)} \right),
\end{equation}
by using
\begin{equation}\label{eq.proposed.diffunc}
\begin{aligned}
f({\bf A},{\bf Z}) = {\bf P}\left[\left({\bf P}^T{\bf A}{\bf Q} - {\bf Q}^T{\bf A}^T{\bf P}\right) \oslash \left({\bf S}{\bf 1}+{\bf 1}{\bf S} \right) \right]{\bf Q}^T \\
+ ({\bf I}-{\bf P}{\bf P}^T) {\bf A} {\bf Q}{\bf S}^{-1}{\bf Q}^T,
\end{aligned}
\end{equation}
where ${\bf P}$, ${\bf Q}$, and ${\bf S}$ are obtained by a SVD ${\bf Z} = {\bf P}{\bf S}{\bf Q}^T$ same as Eq.~\ref{eq.proposed.svd}.
See Appendix.~\ref{sec.app.gradorth} in which we show how to derive Eqs.~\ref{eq.proposed.calcgradZ} and ~\ref{eq.proposed.diffunc}.
Here, $\frac{\partial E}{\partial {\bf C}^{(k)}}$ can be obtained by differentiating Eq.~\ref{eq.proposed.regrecontensor}, resulting as
\begin{equation}\label{eq.proposed.calcgradC}
\frac{\partial E}{\partial {\bf C}^{(k)}} = \frac{1}{N} \sum_{n=1}^{N} \left({\bf \mathcal{X}}_n \prod_{l|l \neq k} \times_l {\bf C}^{(l)} \right)_{(k)}^T \left({\bf \mathcal{W}} \times_1 \frac{\partial E}{\partial {\bf r}_n} - 2 \lambda \overline{{\bf \mathcal{X}}_n} \right)_{(k)},
\end{equation}
where $\frac{\partial E}{\partial {\bf r}_n} = \frac{\partial L(\overline{{\bf \mathcal{X}}_n}, {\bf \theta}; y_n)}{\partial {\bf r}_n}$ is calculated using methods specific to the prediction models, such as the back-propagation algorithm in NNs, by which $\frac{\partial E}{\partial {\bf \theta}'}$ can also be calculated.
Moreover, gradient of ${\bf G}^{(k)}$ can be calculated as
\begin{equation}\label{eq.proposed.calcgradG}
\frac{\partial E}{\partial {\bf G}^{(k)}} = \frac{1}{N} \sum_{n=1}^{N} \left( \overline{{\bf \mathcal{X}}_n} \right)_{(k)}^T \left( \frac{\partial E}{\partial {\bf r}_n} \prod_{l|l \neq k} \odot_1 {\bf G}^{(l)T} \right)_{(1)}.
\end{equation}
The prediction and training algorithm are shown in Algorithm~\ref{alg.training}.

\begin{algorithm}
\caption{\Proposed}\label{alg.training}
\begin{algorithmic}[1]
\Procedure{prediction}{${\bf \mathcal{X}}_n$}
 \State $\overline{{\bf \mathcal{X}}_n} \leftarrow {\bf \mathcal{X}}_n \prod_k \times_k {\bf C}^{(k)}$
 \State $y'_n \leftarrow pred^{\dagger}({\bf r}_n = {\bf \mathcal{W}}_{(1)}^T vec \left( \overline{{\bf \mathcal{X}}_n} \right); {\bf \theta}')$ \Comment{Eqs.~\ref{eq.proposed.model},\ref{eq.proposed.WG}}
 \State \textbf{return} $y'_n$
\EndProcedure
\Procedure{training}{$\{{\bf \mathcal{X}}_n\}_{n=1}^N$, $\{y_n\}_{n=1}^{N}$}
 \State $\{{\bf Z}^{(k)}\}_{k=1}^K$,$\{{\bf G}^{(k)}\}_{k=1}^K$,${\bf \theta}'$ $\leftarrow$ initialize
 \State $\{{\bf C}^{(k)}\} \leftarrow$ orthonormalization( $\{{\bf Z}^{(k)}\}$ ) \Comment{Eqs.~\ref{eq.proposed.svd},\ref{eq.proposed.calcC}}
 \For {$epoch \gets 1$ \textbf{to} $maxepoch$}
  \State set $\{\frac{\partial E}{\partial {\bf Z}^{(k)}}\}, \{\frac{\partial E}{\partial {\bf G}^{(k)}}\}, \frac{\partial E}{\partial {\bf \theta}'}, \{\frac{\partial E}{\partial {\bf r}_n}\}$ to zero
  \For {$n \gets 1$ \textbf{to} $N$}
   \State $y'_n \leftarrow$ {\scriptsize PREDICTION}(${\bf \mathcal{X}}_n$)
   \State $\frac{\partial E}{\partial {\bf \theta}'}, \frac{\partial E}{\partial {\bf r}_n} + \leftarrow$ grad($y_n$,$y'_n$) \Comment{{\it e.g.} back-propagation in NNs}
   \State $\{\frac{\partial E}{\partial {\bf C}^{(k)}}\}, \{\frac{\partial E}{\partial {\bf G}^{(k)}}\} + \leftarrow$ gradCG(${\bf \mathcal{X}}_n$,$\{{\bf C}^{(k)}\}$,$\{{\bf G}^{(k)}\}$,$\frac{\partial E}{\partial {\bf r}_n}$) \Comment{Eqs.~\ref{eq.proposed.calcgradC},\ref{eq.proposed.calcgradG}}
  \EndFor
  \State $\{\frac{\partial E}{\partial {\bf Z}^{(k)}}\} \leftarrow \{f(\frac{\partial E}{\partial {\bf C}^{(k)}}, {\bf Z}^{(k)})\}$ \Comment{Eqs.~\ref{eq.proposed.calcgradZ},\ref{eq.proposed.diffunc}}
  \State $\{{\bf Z}^{(k)}\}, \{{\bf G}^{(k)}\}, {\bf \theta}' \leftarrow$ update($\{\frac{\partial E}{\partial {\bf Z}^{(k)}}\}$,$\{\frac{\partial E}{\partial {\bf G}^{(k)}}\}$,$\frac{\partial E}{\partial {\bf \theta}'}$)
  \State $\{{\bf C}^{(k)}\} \leftarrow$ orthonormalization( $\{{\bf Z}^{(k)}\}$ ) \Comment{Eqs.~\ref{eq.proposed.svd},\ref{eq.proposed.calcC}}
 \EndFor
 \State \textbf{return} $\{{\bf C}^{(k)}\}$,$\{{\bf G}^{(k)}\}$,${\bf \theta}'$
\EndProcedure
\end{algorithmic} 
\end{algorithm}

\subsection{Computational Complexity}

The computation mainly consists of two parts: the projection and the propagation in an NN.
Calculation of the projection scales linearly to the total number of non-zero elements $D$ in the dataset and the size of $\overline{{\bf \mathcal{X}}_n}$, which means the time and space complexity is $O(D \prod_k J_k)$.
The complexity of forward- and back-propagation in an NN is $O(NV)$ time and space, where $V$ is the number of edges in the NN.
Because the number of edges between $\overline{{\bf \mathcal{X}}_n}$ and the first layer is the size of ${\bf \mathcal{W}}$, $V = M \prod_k J_k + V'$ where $V'$ is the number of remaining edges.

\subsection{Understanding Subspace}

\subsubsection{Extraction of Independent Components}

Even if ${\bf C}^{(k)}$ and ${\bf \mathcal{W}}$ are changed to ${\bf C}^{(k)}{\bf R}^{(k)}$ and ${\bf \mathcal{W}} \times_k {\bf R}^{(k)}$ by a rotation transform ${\bf R}^{(k)}$, the value of Eq.~\ref{eq.proposed.regrecontensor} does not change.
We propose to determine ${\bf R}^{(k)}$ so that ${\bf C}^{(k)}{\bf R}^{(k)}$ can be understood as the coefficient of the independent component after regression in modes other than the $k$-th mode.
When the prediction model is a linear model such as ${\bf \mathcal{W}} = {\bf g}^{(1)} \circ \cdots \circ {\bf g}^{(K)}$, ${\bf R}^{(k)}$ can be calculated so that the elements of ${\bf R}^{(k)T}{\bf u}_n^{(k)}$ are independent of each other, where ${\bf u}_n^{(k)} = {\bf \mathcal{X}}_n \prod_{l|l \neq k} \times_l {\bf C}^{(l)}{\bf g}^{(l)} \times_k {\bf C}^{(k)}$.
That is, the column vectors of ${\bf R}^{(k)}$ are the left singular vectors of ${\bf U}^{(k)}$ having the normalized ${\bf u}_n^{(k)}$ as the column vectors.
If the prediction model is a nonlinear model such as an NN, we can use the same strategy as the linear model by learning a {\it linear surrogate model} $\widetilde{y'_n} = \langle \overline{{\bf \mathcal{X}}_n}, \widetilde{\bf \mathcal{W}} \rangle + \widetilde{b}$, where $\widetilde{\bf \mathcal{W}} = \widetilde{\bf g}^{(1)} \circ \cdots \circ \widetilde{\bf g}^{(K)}$ are regression coefficients and $\widetilde{b}$ is a bias.
$\widetilde{\bf \mathcal{W}}$ and $\widetilde{b}$ are calculated so that $\widetilde{y'_n}$ can produce similar results to $y'_n = pred(\overline{{\bf \mathcal{X}}_n})$ by minimizing $\frac{1}{N} \sum_n \| \widetilde{y'_n} - y'_n \|_2^2$.
We can use the alternating least squares (ALS) method~\cite{Kolda09} with which $\widetilde{\bf g}^{(k)}$s are alternately calculated by fixing the coefficients of other modes.

\subsubsection{Local Regression Coefficients}

It might be difficult to understand a strongly nonlinear decision boundary by using a linear surrogate model.
We propose to learn a {\it local linear surrogate model} $\widetilde{y^{'n}_{n'}} = \langle \overline{{\bf \mathcal{X}}_{n'}}, \widetilde{{\bf \mathcal{W}}_n} \rangle + \widetilde{b_n}$, which can be applied only to the data points around the $n$-th point, where $\widetilde{{\bf \mathcal{W}}_n} = \widetilde{{\bf g}_n}^{(1)} \circ \cdots \circ \widetilde{{\bf g}_n}^{(K)}$ are {\it \nameLRC (\LRC)} and $\widetilde{b_n}$ is a bias.
$\widetilde{{\bf \mathcal{W}}_n}$ and $\widetilde{b_n}$ are calculated by minimizing $\frac{1}{N} \sum_{n'} \pi_{n'}^n \| \widetilde{y^{'n}_{n'}} - y'_{n'} \|_2^2$, where $\pi_{n'}^n = exp(-\|\overline{{\bf \mathcal{X}}_{n}} - \overline{{\bf \mathcal{X}}_{n'}}\|_2^2 / \sigma^2)$ is a similarity between $\overline{{\bf \mathcal{X}}_{n}}$ and $\overline{{\bf \mathcal{X}}_{n'}}$ and $\sigma$ is a parameter.
This is almost the same as the local interpretable model proposed by Ribeiro et al.~\cite{Ribeiro16} except that we apply it to the projected tensor data.
We can calculate the parameters by using the ALS method, the same as the linear surrogate model.
\LRC can be observed using ${\bf R}^{(k)}$ calculated under the local linear surrogate model.
For $K=1$, we should observe ${\bf R}^{(1)T}\widetilde{{\bf g}_n}^{(1)}$.
For $K>1$, \LRC should be projected in modes other than the $k$-th mode, resulting in ${\bf R}^{(k)T} \widehat{{\bf g}_n}^{(k)}$ where $\widehat{{\bf g}_n}^{(k)} = \widetilde{{\bf g}_n}^{(k)} \prod_{l|l \neq k} \langle \widetilde{{\bf g}_n}^{(l)}, \widetilde{{\bf g}}^{(l)} \rangle$.
Also, we can understand \LRC as ${\bf C}^{(k)} \widehat{{\bf g}_n}^{(k)}$ in the language of the original variable.

\section{Empirical Results}
\label{sec.exp}
\subsection{Evaluation Settings}

\subsubsection{Datasets}

\begin{table}[tb]
\caption{Summary of datasets. $K$: order of the dataset. $I$: number of variables. $N$: number of samples. $b$: mini batch size.\label{tbl.exp.dataset}}
\centering
\begin{tabular*}{\columnwidth}{@{\extracolsep{\fill}}lllrrrl}
\hline
$K$ & dataset & cite & $I$ & $N$ & $b$ & task \\
\hline
\hline
1 & \Spiral (artificial) && 100 & 220 & 20 & classification \\
& \Randomone (artificial) && 1,000 & 100 & 32 & (2 class) \\
& \Connectionist~\footnotemark[1] & \cite{Dua:2019} & 60 & 208 & 64 & \\
& \quality~\footnotemark[1] & \cite{Dua:2019,fernandes_transfer_2017} & 62 & 287 &64 & \\
& \Alizadeh~\footnotemark[1] & \cite{Dua:2019,alizadehsani_data_2013} & 57 & 303 & 64 & \\
& \Sports~\footnotemark[1] & \cite{Dua:2019,Hajj2018ASC} & 59 & 1,000 & 256 & \\
& \SECOM & \cite{Dua:2019} & 590 & 1,567 & 256 & \\
& \Ozone~\footnotemark[1] & \cite{Dua:2019} & 72 & 1,848 & 256 & \\
& \Spambase & \cite{Dua:2019} & 57 & 4,601 & 1,024 & \\
& \theo~\footnotemark[1] & \cite{Dua:2019,Bridge14} & 51 & 6,118 & 1,024 & \\
& \Epileptic~\footnotemark[1] & \cite{Dua:2019,Andrzejak02} & 178 & 11,500 & 2,048 & \\
\cline{2-7}
& \Iris & \cite{Dua:2019} & 4 & 150 & 32 & classification \\
& \Wine & \cite{Dua:2019} & 13 & 178 & 32 & (3 class) \\
\hline
2 & \GeneExpression & \cite{Shimamura_11} & 13,508 $\times$ 1,732 & 762 & 128 & regression \\
\cline{2-7}
& \Randomtwo (artificial) && 1,000 $\times$ 1,000 & 100 & 32 & classification \\
\cline{1-6}
3 & \Randomthree (artificial) && 1,000 $\times$ 1,000 $\times$ 1,000 & 100 & 32 & (2 class) \\
\hline
\end{tabular*}
\end{table}
\footnotetext[1]{\Connectionist: \nameConnectionist, \quality: \namequality, \Alizadeh: \nameAlizadeh, \Sports: \nameSports, \Ozone: \nameOzone, \theo: \nametheo, \Epileptic: \nameEpileptic.}

Table~\ref{tbl.exp.dataset} shows the summary of the datasets used in this study.
All the vector datasets ($K=1$), except the artificial datasets, are downloaded from the website of {\it UCI Machine Learning Repository}\footnote[2]{https://archive.ics.uci.edu/ml/index.php}~\cite{Dua:2019}.
We obtain \GeneExpression from a website\footnote[3]{http://bonsai.hgc.jp/\~{}shima/NetworkProfiler/}~\cite{Shimamura_11}.
We normalize all the variables of the vector datasets by making the average values zero and the standard deviations one.

\subsubsection{Comparison Methods}

We compare \Proposed with methods that can be applied to the \NTR problem, which linearly project data points into low-dimensional subspace and predict scalar responses using prediction models with strong nonlinearity.
As in \Proposed, NNs are used for the prediction models.
The compared methods for vector datasets use PCA and LDA for linear projection, and the methods combined with NNs are called {\it \PCANN} and {\it \LDANN}, respectively.
For tensor datasets ($K>1$), we compare \Proposed with a method of applying NNs to tensors projected to the low-dimensional subspace by Tucker decomposition~\cite{Kolda09}, {\it i.e.}, core tensors.
This method is called {\it \TuckerNN}.

In all the experiments other than evaluation using \Spiral and scalability tests, we select parameters of each method based on the average accuracies of $5$ trials of $10$-fold cross validation.
For \Spiral, we use the average test accuracies of $10$ trials.
The number of hidden layers of NNs are chosen from zero (linear/softmax regression) to $4$, with 10 neurons in each layer.
We use ReLU as activation function.
Also, we use softmax cross entropy loss for classification and squared error loss for regression.
The best $\lambda$s for \Proposed are chosen from $\{0.00001, 0.0001, 0.001, 0.01, 0.1, 1, 10\}$.
The number of epochs in SGD is determined based on sufficient convergence, {\it i.e.}, $2,000$ for \Spiral, $1,000$ for the datasets of $2$-class classification task other than artificial datasets, $2,000$ (\Proposed) and $500$ (\PCANN, \LDANN) for \Iris and \Wine, $50$ (\Proposed) and $200$ (\TuckerNN) for \GeneExpression, and $100$ for \Randomone, \Randomtwo, and \Randomthree.
We use Adam optimization algorithm with learning rate of $0.001$, whereas we use $0.01$ for \GeneExpression.

\Proposed, \PCANN, \LDANN, and \TuckerNN are implemented in Python 3.7 using PyTorch 1.4\footnote[4]{https://pytorch.org/}.
All the experiments are conducted on an Intel(R) Xeon(R) Gold 6130 CPU 2.10GHz with 32GB of memory, running Linux.

\subsection{Evaluation using Artificial Data}

\subsubsection{Data Generation}

We create artificial data \Spiral in which essential decision boundaries are carefully hidden in high-dimensional space.
First, we generate a two-dimensional distribution in which the data points of the two classes are distributed in a two-dimensional spiral, as shown in Fig.~\ref{fig.intro.galaxy}(a).
Specifically, we randomly select $x$ from a uniform distribution of $0.0 < x < 3.5 \pi$ and calculate the two-dimensional coordinates, {\it i.e.}, $(x \sin(x), x \cos(x))$ for the $100$ samples of the first class and $(x \sin(x + \pi), x \cos(x + \pi))$ for the $100$ samples of the second class.
We also add $10$ samples to each class as {\it noise samples}, with the two-dimensional coordinates of $(x_1, x_2)$ selected from the uniform distribution of $-10.0 < x_1, x_2 < 10.0$.
We normalize the standard deviation of the coordinates in each dimension of these $220$ samples to $1.0$.
We also create the coordinates of an additional one dimension as a {\it major noise dimension} from a normal distribution with the standard deviation of $1.0$, which cannot be distinguished from the dimensions with spiral distribution with PCA because of the same variance.
In addition, we add the coordinates of $97$ dimensions as {\it minor noise dimensions} from a normal distribution with the standard deviation of $0.1$, which should cause model overfitting when the regularization of reconstruction errors does not sufficiently work (Fig.~\ref{fig.intro.galaxy}(b)).
Finally, the spiral distribution is hidden in the $100$-dimensional space by transforming the data using a randomly generated rotation transform.
As a result, we generate artificial dataset containing $220$ samples with $100$-dimensional coordinates.
We generate the two sets of samples by selecting coordinates independently, {\it i.e.}, training dataset and test dataset, using the procedure described above with the same rotation transform.

\subsubsection{Accuracy Evaluation and Visualization}

\begin{figure}[ttt]
\begin{center}
\includegraphics[width=\columnwidth]{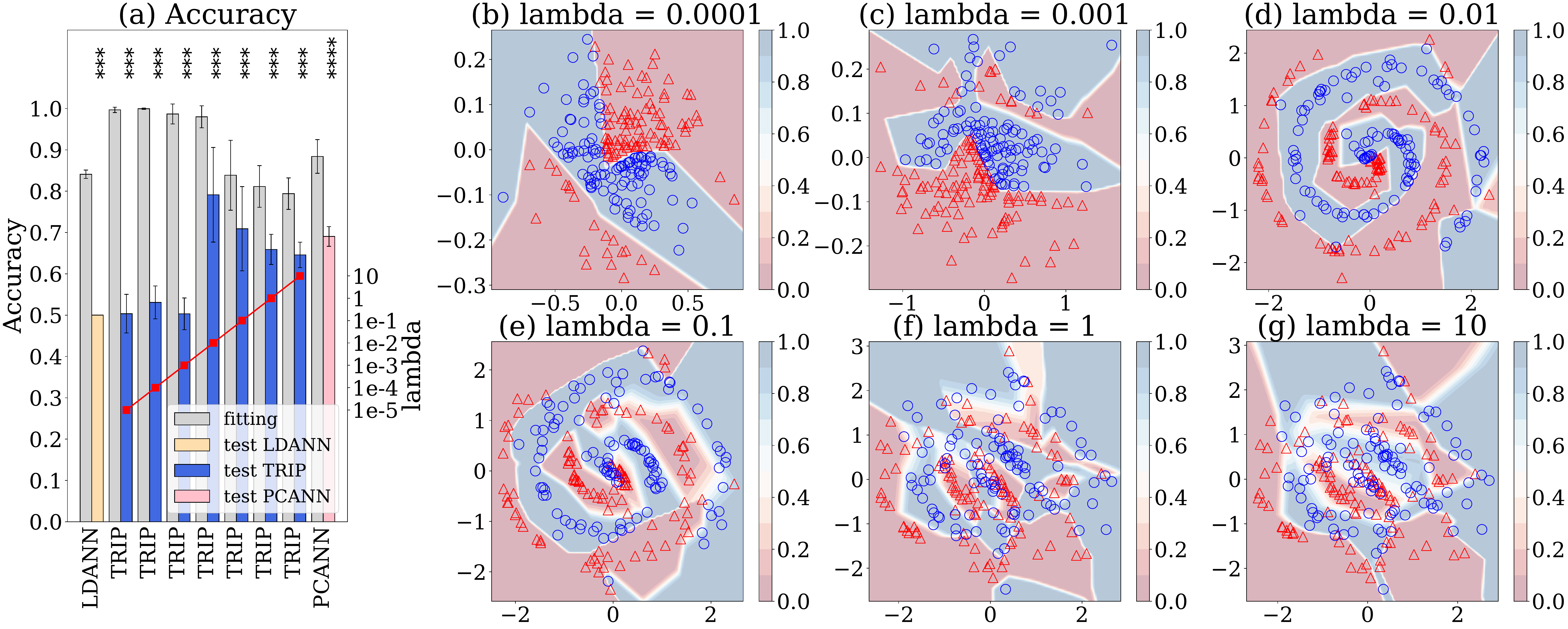}
\caption{(a) Fitting and test accuracies of \Spiral for several values of $\lambda$ using $3$ hidden layers of NNs applied to the $2$-dimensional subspace. Line plot shows the value of $\lambda$. The number of hidden layers of NNs is indicated by the number of asterisks ($\ast$).
(b-g) Distribution of the projected data points of the training data of \Spiral for several values of $\lambda$ using $3$ hidden layers of NNs. Contours represent the softmax value for class $1$. Blue circles: class $1$, red triangles: class $2$.
\label{fig.exp.galaxy}}
\end{center}
\end{figure}

Because we know that the essential decision boundaries have the two-dimensional distribution with spiral shape, we conduct the experiments using two projection vectors, {\it i.e.}, $J_1 = 2$. 
The highest test accuracies are achieved with an NN with $3$ hidden layers and $\lambda = 0.01$.
Figure~\ref{fig.exp.galaxy}(a) shows the fitting accuracies on the training dataset and the test accuracies on the test dataset with the NN with $3$ hidden layers for several values of $\lambda$, along with the results of \LDANN and \PCANN.
Overall, lower $\lambda$ cause overfitting in which fitting accuracies are high whereas test accuracies are low.
On the other hand, higher $\lambda$ can balance between the fitting accuracies and the test accuracies, which indicates that the reconstruction errors work as regularization functions.
The result of \LDANN is similar to those of low $\lambda$, and the result of \PCANN is similar to those of high $\lambda$.

Figure.~\ref{fig.exp.galaxy}(b-g) plot the projected data points of the training dataset with the NN with $3$ hidden layers for several values of $\lambda$.
All the plots are chosen from $10$ trials because they reduce the objective function the most.
We can recognize the spiral boundaries for the $\lambda$ of $0.01$ and $0.1$ (Fig.~\ref{fig.exp.galaxy}(d)(e)).
For the $\lambda$ less than $0.01$, the samples of the two classes are well separated (Fig.~\ref{fig.exp.galaxy}(b)(c)); however, the test accuracies are very low (Fig.~\ref{fig.exp.galaxy}(a)), suggesting the models overfit to the training dataset for these $\lambda$.
When we use $\lambda$ more than $0.1$, the samples of the two classes are not clearly separated (Fig.~\ref{fig.exp.galaxy}(f)(g)), the distributions of which are similar to that of \PCANN (Fig.~\ref{fig.intro.galaxy}(e)).
The results on $\lambda$ more than $0.1$ suggest that the projection vectors are learned in an almost unsupervised manner.
Thus, by using the appropriate $\lambda$, the reconstruction error works properly as regularization, and \Proposed extracts the subspace that captures the inherently important features for separation in high-dimensional data.

\subsection{Evaluation by Real-World Data}

\subsubsection{Evaluation of Accuracies}

\begin{figure}[ttt]
\begin{center}
\includegraphics[width=\columnwidth]{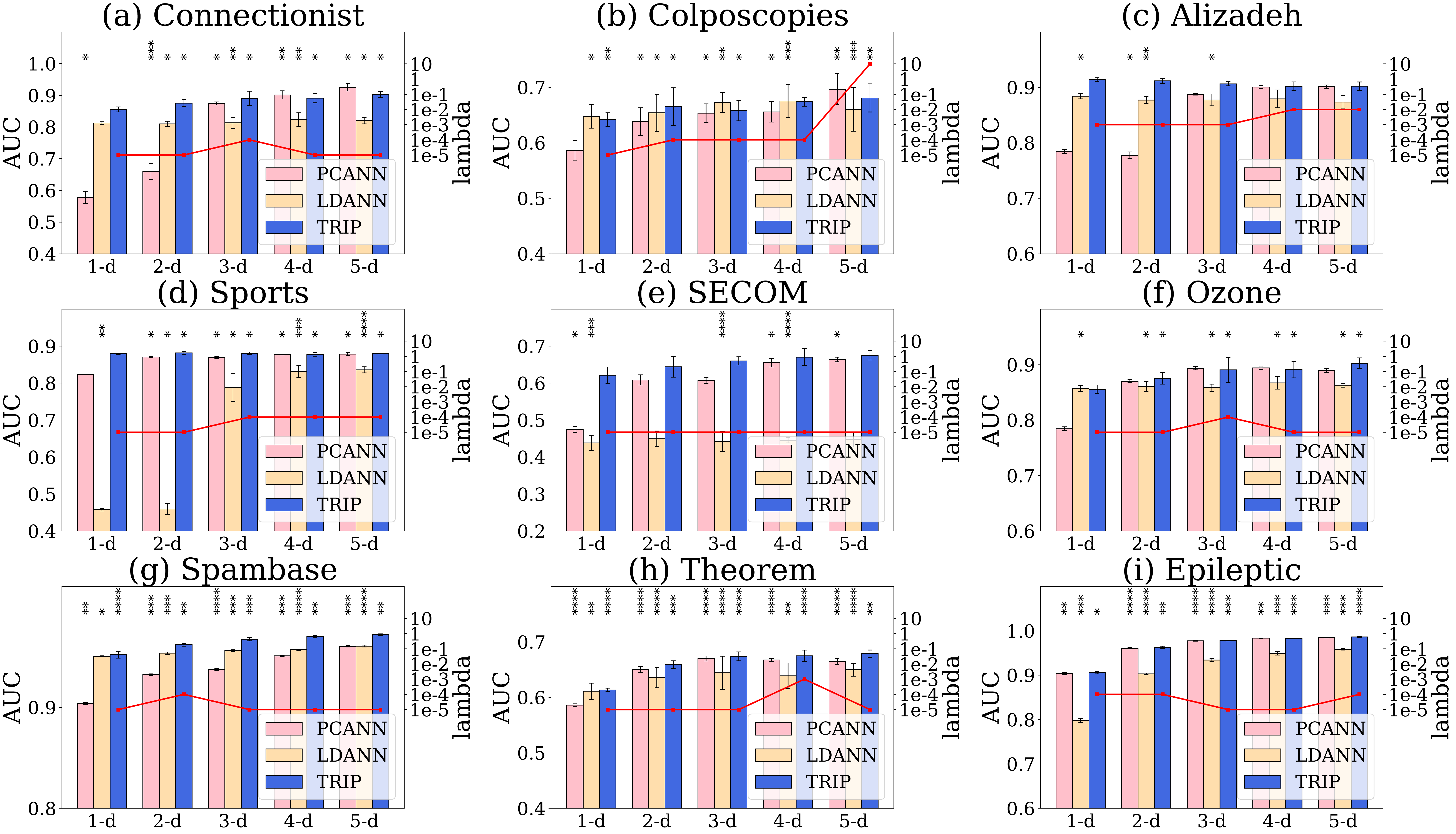}
\caption{
AUCs when projected into $1$- to $5$-dimensional subspace using \PCANN, \LDANN, and \Proposed on $2$-class classification tasks. Line plot shows the value of $\lambda$. The number of hidden layers of NNs is indicated by the number of asterisks ($\ast$).\label{fig.exp.vector_PCA_NTR3_test}}
\end{center}
\end{figure}

To quantitatively show the effectiveness of \Proposed, the area under the curves (AUCs) calculated by $10$-fold cross validations for various $2$-class classification tasks are shown in Fig.~\ref{fig.exp.vector_PCA_NTR3_test}.
\Proposed achieves higher accuracies than \PCANN and \LDANN in most of the datasets, especially when the data points are projected into $1$- or $2$- dimensions.
Also, datasets with many samples require the NNs with many hidden layers (Fig.~\ref{fig.exp.vector_PCA_NTR3_test}(g)(h)(i)).
This indicates that the decision boundaries of these datasets contain relatively strong nonlinearities.
These results suggest that \Proposed is good at linear projection while maintaining high prediction accuracy in a low-dimensional subspace that is easy for humans to understand, even when the decision boundary has high nonlinearity.

\subsubsection{Visualizing Subspace}

\begin{figure}[ttt]
\begin{center}
\includegraphics[width=\columnwidth]{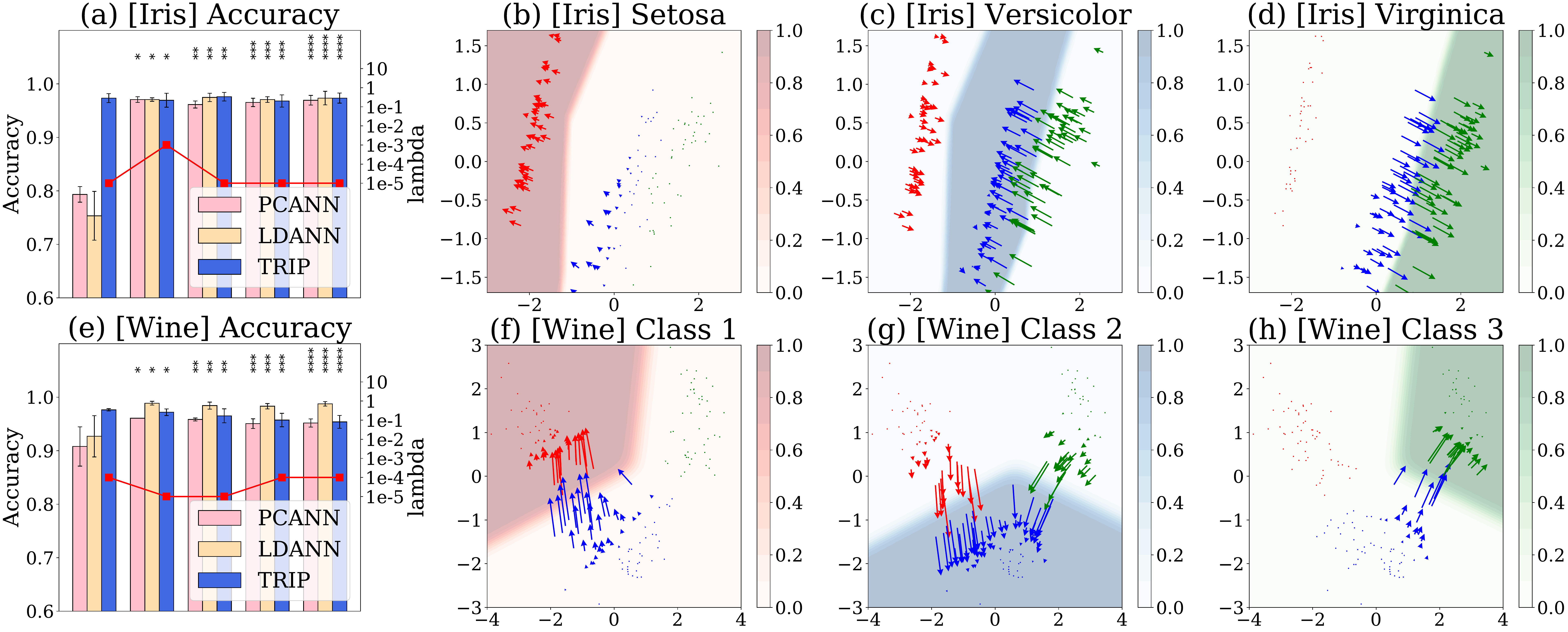}
\caption{(a)(e) Accuracies of \PCANN, \LDANN, and \Proposed on \Iris and \Wine.
We use NNs with $0$ to $4$ hidden layers indicated by the number of asterisks ($\ast$), applied to the $2$-dimensional subspace.
Line plot shows the value of $\lambda$.
(b)(c)(d)(f)(g)(h) Softmax value of \Proposed for each class of \Iris and \Wine is represented as contours.
Arrows indicate \LRC for the data points calculated using $\sigma=1.0$.
Red, blue, and green arrows are Setosa, Versicolor, and Virginica for \Iris, and Class $1$, $2$, and $3$ for \Wine, respectively.
\label{fig.exp.NTR3}}
\end{center}
\end{figure}

We show how to understand the reason for the prediction using two datasets of $3$-class classification, \Iris and \Wine.
We use two-dimensional subspace, {\it i.e.}, $J_1 = 2$, to understand the results in the two-dimensional plots.
Figure~\ref{fig.exp.NTR3}(a)(e) shows the accuracies of \PCANN, \LDANN, and \Proposed using NNs with $0$ to $4$ hidden layers.
Though the best accuracies are almost same among the methods, \Proposed achieves high accuracies even with the model without hidden layers of the NN.
We choose the number of hidden layers of the NN as $2$ for \Iris and $0$ for \Wine, and $\lambda = 0.00001$ for \Iris and $\lambda = 0.0001$ for \Wine, by which \Proposed achieves the highest accuracies.

We show the softmax value of \Proposed as contours for each class of \Iris (Fig.~\ref{fig.exp.NTR3} (b)(c)(d)) and \Wine (Fig.~\ref{fig.exp.NTR3} (f)(g)(h)).
The two-axis are rotated as described in Section~\ref{sec.proposed}.
We also represent \LRC of each data point with an arrow.
In \Iris, data points of {\it Versicolor} and {\it Virginica} are very close, and \LRC of most of these data points have almost same direction (Fig.~\ref{fig.exp.NTR3} (c)(d)).
These \LRC can be interpreted in the language of the original variables, {\it i.e.}, ${\bf C}^{(k)} \widehat{{\bf g}_n}^{(k)}$.
Roughly speaking, the direction of \LRC means Virginica has larger petal and smaller sepal than Versicolor.
In \Wine, it can be observed that large \LRC of data points near the decision boundary separating the class $1$ and $2$ (Fig.~\ref{fig.exp.NTR3} (f)) or the class $2$ and $3$ (Fig.~\ref{fig.exp.NTR3} (g)) have almost same direction.
In short, in order for a class $2$ wine close to the decision boundary to be a class $1$ or class $3$ wine, at least the value of {\it Color intensity}, {\it Proline}, {\it Alcohol}, and {\it Ash} must be higher and the value of {\it Hue} must be lower.
In addition, the value of {\it Alcalinity of ash} needs to be smaller to be a class $1$ wine, whereas the value of {\it Flavanoids} must be smaller to be a class $3$ wine.
Thus, \LRC makes it easy to understand the reason for prediction, in a linearly projected low-dimensional subspace.

\subsubsection{Results on Higher-Order Dataset}

\begin{figure}[ttt]
\begin{center}
\includegraphics[width=\columnwidth]{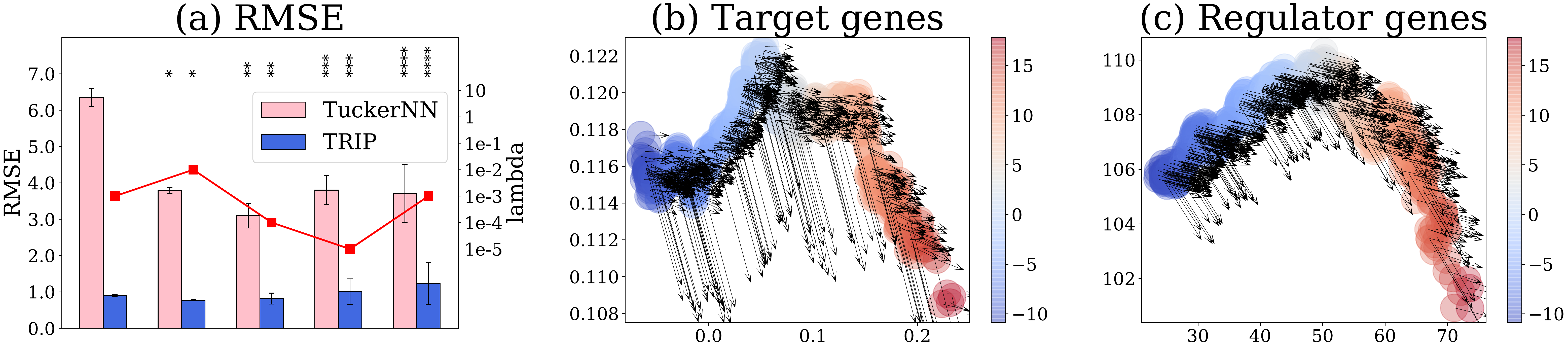}
\caption{(a) RMSE of \TuckerNN and \Proposed on \GeneExpression.
We use NNs with $0$ to $4$ hidden layers indicated by the number of asterisks ($\ast$), applied to the $2 \times 2$ dimensional subspace.
Line plot shows the value of $\lambda$.
(b)(c) Values of data points in \GeneExpression projected on first (horizontal) and second (vertical) axes by \Proposed.
Arrows indicate \LRC for the data points calculated using $\sigma=1.0$.
Color indicates the predicted EMT values.
\label{fig.exp.gene_NTR3}}
\end{center}
\end{figure}

As a higher-order dataset, we use \GeneExpression in which each data point is a matrix, {\it i.e.}, the second order tensor.
This dataset describes {\it \nameVC (\VC)} between $13,508$ {\it \nameTG (\TG)} and $1,732$ {\it \nameRG (\RG)} in $762$ cell lines~\cite{Shimamura_11}.
\VC are computed based on the expression levels of genes, where the strengths of relationship between genes is allowed to vary with {\it epithelial-mesenchymal transition (EMT)-related modulator value} (hereinafter called {\it EMT value}), which describe cell lines as epithelial-like or mesenchymal-like.
We set \VC with the absolute values less than $0.3$ as zero, resulting in $2,118$ non-zero elements on average.
We focus on the regression task to predict the EMT values.
We project each data point of the cell line to the subspace of the size of $2 \times 2$, {\it i.e.}, $J_1=2$ and $J_2=2$.
Figure~\ref{fig.exp.gene_NTR3}(a) shows the root mean squared error (RMSE) of \Proposed and \TuckerNN for predicting EMT value using NNs with different numbers of hidden layers.
\Proposed shows much better accuracy than \TuckerNN, even when using NNs with no or only one hidden layer.
We choose the model of one hidden layer of the NN and $\lambda=0.01$ as the best parameters.
Figure~\ref{fig.exp.gene_NTR3}(b)(c) plots the data points projected to the two-axis.
As described in Section~\ref{sec.proposed}, the two axes of each mode are rotated and \LRC of every data points indicated by the arrows are calculated.
Note that the predicted values have a high Pearson's correlation coefficient with \TG and \RG on the first axis ($1.000$ and $0.999$), while low on the second axis ($-0.002$ and $-0.040$). 
%Interestingly, value on the second axes is partially correlated with value of the first axes, that is, there is positive correlation between them for lower value of the first axes, and negative correlation for higher value.
Interestingly, there is positive correlation between value of the first axes and the second axes for lower value of the first axes, and negative correlation for higher value.
Moreover, a mixture of \LRC in two different directions is observed throughout the data.
This suggests that there are at least two gene groups in both \TG and \RG, which interact in a complex manner.

We show \TG and \RG that have the highest absolute scores of ${\bf C}^{(k)}{\bf R}^{(k)}$ on the two axes in Table~\ref{tbl.exp.genes_list}.
Note that we only consider absolute scores because the sign of the scores becomes unstable since the projection is performed by the outer product of the projection vectors of each mode.
We analyze the top-$50$ genes by the bioinformatics tool DAVID~\footnote[5]{https://david.ncifcrf.gov/}~\cite{huang_systematic_2009}, which provides functional annotation clustering of genes and their biological interpretation.
It can be seen through annotation summary results of {\it Tissue expression (t\_e)} that there is a clear difference between the axes; the {\it Platelet}, {\it Thyroid}, and {\it Thymus} are derived as annotations related to the first axis of \TG, whereas annotations {\it Ovary} and {\it Eye} are extracted for the second axis of \TG.
On the other hand, the results for \RG include only {\it Epithelium} in the first axis, whereas {\it Muscle}, {\it Pancreatic cancer}, {\it Colon cancer}, {\it Uterus}, {\it Colon adenocarcinoma}, {\it Colon tumor}, and {\it Peripheral blood lymphocyte} in the second axis.
These observations suggest that there are other biomechanisms involved in various ways with EMT, independent of the relationships between \TG and \RG extracted in the first axis that can directly predict the EMT value.
We hope that these results will be analyzed more deeply by biological researchers, leading to the discovery of new biological mechanisms.

\begin{table}[tb]
\begin{center}
\caption{Genes with high absolute scores. PL: {\it Platelet}, TR: {\it Thyroid}, TM: {\it Thymus}, OV: {\it Ovary}, EY: {\it Eye}, EP: {\it Epithelium}, MS: {\it Muscle}, UT: {\it Uterus}, and CT: {\it Colon tumor}.\label{tbl.exp.genes_list}}
\centering
\scriptsize
\begin{tabular*}{\columnwidth}{@{\extracolsep{\fill}}ccccccccccccccc}
\hline
\multicolumn{3}{c}{Target \# 1} && \multicolumn{3}{c}{Target \# 2} && \multicolumn{3}{c}{Regulator \# 1} && \multicolumn{3}{c}{Regulator \# 2} \\
\cline{1-3} \cline{5-7} \cline{9-11} \cline{13-15}
score & gene & t\_e && score & gene & t\_e && score & gene & t\_e && score & gene & t\_e \\
\hline
\hline
0.084 & C17orf39 &&& 0.025 & GTF2H3 & EY && 0.128 & NCOR1 & EP && 0.097 & DIP2A & UT \\
0.083 & SLC6A8 &&& 0.025 & KIAA*~\footnotemark[6] &&& 0.118 & IKZF1 &&& 0.035 & m451~\footnotemark[6] &\\
0.072 & MPP1 & TM && 0.024 & TIGD1L &&& 0.110 & LYL1 &&& 0.034 & SAP*~\footnotemark[6] & MS,UT \\
0.067 & PLOD3 &&& 0.023 & TCF7 &&& 0.109 & DIP2A &&& 0.032 & HSF1 & MS \\
0.062 & CHKA &&& 0.022 & LOC*~\footnotemark[6] &&& 0.090 & AIP &&& 0.032 & NAT8~\footnotemark[6] & CT \\
0.058 & SASH1 & TR && 0.022 & PRR11 & EY && 0.089 & SOX10 &&& 0.032 & m422a*~\footnotemark[6] &\\
0.055 & AFTPH &&& 0.022 & ERN2 &&& 0.088 & m223~\footnotemark[6] &&& 0.031 & LMCD1 &\\
0.054 & GSPT1 &&& 0.022 & MBTD1 & EY && 0.086 & ELF3 &&& 0.030 & TAF12 &\\
0.053 & MSX2 &&& 0.021 & MTRF1L & EY && 0.086 & TTF1 & EP && 0.030 & LBA1 &\\
0.051 & COASY & PL && 0.021 & UBQLN4 & OV && 0.081 & CREB3 &&& 0.030 & NOTCH2 &\\
\hline
\end{tabular*}
%}
\end{center}
\end{table}

\subsection{Evaluation of Scalability}

\begin{figure}[ttt]
\begin{center}
\includegraphics[width=3.2in]{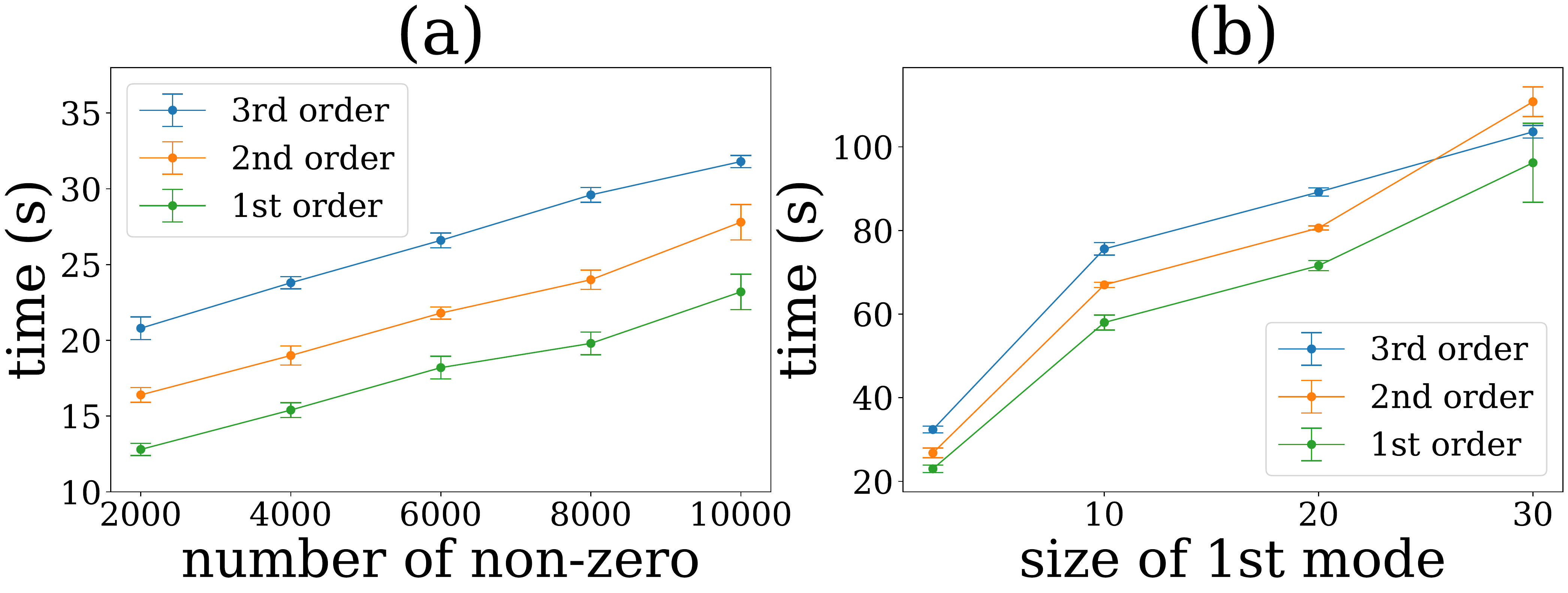}
\caption{Scalability of \Proposed. The training time of \Proposed scales linearly to the number of non-zero elements (a) and the size of projected data (b).\label{fig.exp.scalability}}
\end{center}
\end{figure}

To empirically prove the scalability of \Proposed, we create artificial datasets of first-, second-, and third-order tensors, {\it i.e.}, \Randomone, \Randomtwo, \Randomthree in Table~\ref{tbl.exp.dataset}, respectively, by randomly selecting non-zero elements.
We only show the results of the model using two hidden layers of the NN and $\lambda=0.01$; however, we obtained almost the same results for other settings (not shown).
Figure~\ref{fig.exp.scalability} (a) shows that the training time of \Proposed increases linearly to the number of non-zero element of the datasets.
Figure~\ref{fig.exp.scalability} (b) also shows that the training time increases linearly to the size of the projected tensor, where we change the size of one dimension of the projected tensor while fixing the size of the other dimensions as $2$.
Note that the number of non-zero elements are fixed to $100$.
\footnotetext[6]{KIAA*: KIAA0894, LOC*: LOC100272216, m223: hsa-miR-223, m451: hsa-miR-451, SAP*: SAP30BP, NAT8: NAT8 /// NAT8B, m422a*: hsa-miR-422a*.}

\section{Conclusion}
\label{sec.concl}
We proposed a method, \Proposed, for searching linear projection that maximizes prediction accuracy while retaining the original data variance as much as possible.
Experiments using artificial data showed that \Proposed correctly finds decision boundaries among many variables even with strong nonlinearity.
Experiments with real-world data, including higher-order tensors, also showed that \Proposed makes it possible to find various events behind the data, using human-readable low-dimensional subspace.
Moreover, we empirically showed that \Proposed scales linearly to the number of non-zero elements and the size of projected tensor.

%\bibliographystyle{unsrt}
%\bibliography{BIB/paper}

\appendix

\section{Calculating Gradient in Orthonormal Conditions}
\label{sec.app.gradorth}
How we derive Eqs.~\ref{eq.proposed.calcgradZ} and ~\ref{eq.proposed.diffunc} is shown below.
For readability, superscripts are omitted.
The ${\bf C}$ defined by Eqs.~\ref{eq.proposed.svd} and ~\ref{eq.proposed.calcC} is calculated as a matrix that satisfies 
\begin{equation}\label{eq.ZC}
{\bf Z}{\bf C}^T = {\bf C}{\bf Z}^T, \quad {\bf C}^T{\bf C} = {\bf I},
\end{equation}
because the former equation of Eqs.~\ref{eq.ZC} can be written as ${\bf P}{\bf S}{\bf Q}^T{\bf C}^T = {\bf C}{\bf Q}{\bf S}{\bf P}^T$ by using Eq.~\ref{eq.proposed.svd} and this is regarded as a SVD, where the diagonal components of ${\bf S}$ are singular values and the column vectors of ${\bf P} = {\bf C}{\bf Q}$ are left singular vectors.
Thus ${\bf C} = {\bf P}{\bf Q}^T$ can be obtained.
By differentiating both sides of both equations of Eqs.~\ref{eq.ZC} with the $(i,j)$-th component of ${\bf Z}$, we can obtain
\begin{equation}\label{eq.ZCdif}
\frac{\partial {\bf Z}}{\partial z_{ij}}{\bf C}^T + {\bf Z} \frac{\partial {\bf C}^T}{\partial z_{ij}} = \frac{\partial {\bf C}}{\partial z_{ij}}{\bf Z}^T + {\bf C} \frac{\partial {\bf Z}^T}{\partial z_{ij}}, \quad \frac{\partial {\bf C}}{\partial z_{ij}}^T {\bf C} + {\bf C}^T \frac{\partial {\bf C}}{\partial z_{ij}} = {\bf 0}.
\end{equation}
The $\frac{\partial {\bf C}}{\partial z_{ij}}$ that satisfies Eqs.~\ref{eq.ZCdif} is $\frac{\partial {\bf C}}{\partial z_{ij}} = f(\frac{\partial {\bf Z}}{\partial z_{ij}},{\bf Z})$ by using Eq.~\ref{eq.proposed.diffunc}.
If we use the property of $\langle {\bf A}{\bf B}, {\bf C} \rangle = \langle {\bf A}, {\bf C}{\bf B}^T \rangle$, we obtain
\begin{equation}\label{eq.ZCdifPQrev}
\frac{\partial E}{\partial z_{ij}} = \left\langle \frac{\partial E}{\partial {\bf C}}, \frac{\partial {\bf C}}{\partial z_{ij}} \right\rangle = \left\langle \frac{\partial E}{\partial {\bf C}}, f\left(\frac{\partial {\bf Z}}{\partial z_{ij}},{\bf Z}\right) \right\rangle = \left\langle f\left(\frac{\partial E}{\partial {\bf C}},{\bf Z}\right), \frac{\partial {\bf Z}}{\partial z_{ij}} \right\rangle.
\end{equation}
The $\frac{\partial {\bf Z}}{\partial z_{ij}}$ is $1$ for the $(i,j)$-th component otherwise $0$.
Thus, we can derive Eq.~\ref{eq.proposed.calcgradZ}.

\end{document}